# Texture Enhancement via High-Resolution Style Transfer for Single-Image Super-Resolution

Il Jun Ahn[†] and Woo Hyun Nam[†]*

**Abstract** — *Recently, various deep-neural-network (DNN)-based approaches have been proposed for single-image super-resolution (SISR). Despite their promising results on major structure regions such as edges and lines, they still suffer from limited performance on texture regions that consist of very complex and fine patterns. This is because, during the acquisition of a low-resolution (LR) image via down-sampling, these regions lose most of the high frequency information necessary to represent the texture details. In this paper, we present a novel texture enhancement framework for SISR to effectively improve the spatial resolution in the texture regions as well as edges and lines. We call our method, high-resolution (HR) style transfer algorithm. Our framework consists of three steps: (i) generate an initial HR image from an interpolated LR image via an SISR algorithm, (ii) generate an HR style image from the initial HR image via down-scaling and tiling, and (iii) combine the HR style image with the initial HR image via a customized style transfer algorithm. Here, the HR style image is obtained by down-scaling the initial HR image and then repetitively tiling it into an image of the same size as the HR image. This down-scaling and tiling process comes from the idea that texture regions are often composed of small regions that similar in appearance albeit sometimes different in scale. This process creates an HR style image that is rich in details, which can be used to restore high-frequency texture details back into the initial HR image via the style transfer algorithm. Experimental results on a number of texture datasets show that our proposed HR style transfer algorithm provides more visually pleasing results compared with competitive methods.[1]*

**Index Terms** — Single-image super-resolution, texture enhancement, high-resolution style, style transfer, down-scaling and tiling.

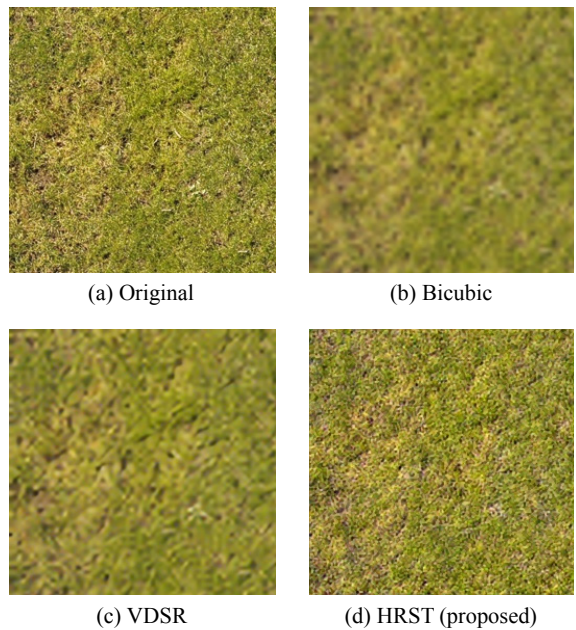

Fig. 1. Our high-resolution style transfer (HRST) based SISR method compares favorably with a representative related work [19] on texture region (up-sampling factor is 4.).

## I. INTRODUCTION

The aim of single-image super-resolution (SISR) algorithm is to recover a high-resolution (HR) image from a single low-resolution (LR) image [1]. Although the SISR problem inherently ill-posed, many valuable algorithms have been presented for computer vision and image processing applications such as surveillance imaging, medical imaging, or ultra-high-definition (UHD) image generation where more image details are required. Early methods include simple and fast interpolation-based scheme with bicubic or Lanzcos filter [2]. For better performance, more advanced schemes using statistical image priors [3]-[7] or internal patch recurrence [8], [9] were also introduced.

Meanwhile, sophisticated machine learning based schemes have been widely used to learn the relationship from LR to HR patches. Neighborhood embedding approaches [10], [11] up-sample a given LR image patch by finding similar LR training patches in a low dimensional manifold and combining their corresponding HR patches for reconstruction. Sparse-coding (or dictionary learning) approaches [12]-[14] use a learned compact dictionary on the basis that natural patches can be represented using sparse activations of dictionary atoms. Random forests approaches [15] directly formulate SISR as a regression problem, which can avoid complex and time-consuming training of a sparse dictionary.

Recently, various deep learning-based approaches via convolutional neural networks (CNN) were proposed with excellent performance. Dong et al. [16], [17] showed that CNN could be successfully applied for SISR. This CNN method, which we call SRCNN, used a three layer convolutional network and trained in an end-to-end manner to learn a mapping from interpolated LR image to original HR

[1] The authors thank to Mr. Kiheum Cho, Dr. Yongsup Park, and Ms. Tammy Lee in the Digital Media & Communications R&D Center, Samsung Electronics, Seoul, Korea for the helpful discussions and collaboration.

Il Jun Ahn and Woo Hyun Nam are with the Digital Media & Communications R&D Center, Samsung Electronics, Seoul, Korea. (*Corresponding author: W. H. Nam, E-mail: woohyun.nam@samsung.com)

[†]Both authors contributed equally to this work.

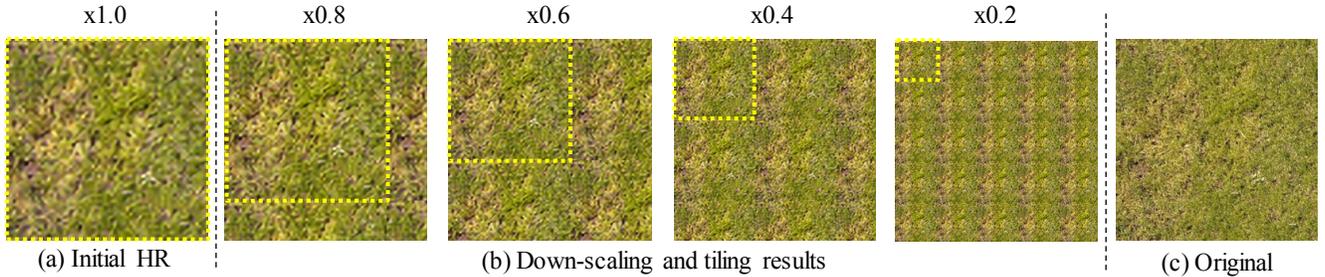

Fig. 2. (a) Initial HR image enhanced from the interpolated LR image via a SISR, (b) various results obtained by down-scaling and tiling the initial HR image based on scaling factors ranging from 0.2 to 0.8, and (c) original HR image. In terms of texture detail representation, the result with scaling factor 0.4 is most similar to the original image. The yellow boxes in (b) denote down-scaled ones of initial HR image.

image. To further improve the performance on both accuracy and speed, the authors extended their work to enable the network to learn the mapping from LR to HR image directly, rather than from the interpolated LR image [18]. Since up-sampling is only performed in the last layer of the network, the method can avoid expensive computations in the HR dimension.

Kim et al. [19] presented a highly performant architecture which consists of very deep convolutional network of 20 layers. Since the deep networks lead to enlargement of receptive fields that can take a large image context into account, the method achieved state-of-the art performance with a large margin. They also presented a novel residual learning approach and showed that it is more favorable in training the deep layers than a non-residual based one. Meanwhile, to reduce the number of convolutional parameters while keeping the large receptive fields, the authors proposed a different architecture based on deeply recursive convolutional network [20], which showed comparable performance to [19].

Despite the promising results of the recent SISR algorithms, compared with original HR image, they still show overly smoothed results and/or lack of high-frequency details, especially on texture regions (see Fig. 1(a)-(c).). In an attempt to resolve this problem, Johnson et al. [21] suggested using perceptual loss, instead of conventional mean squared reconstruction error (MSE), and Ledig et al. [22] proposed a notable SR framework combined with generative adversarial network (SRGAN). Even though quantitative performance of these methods, such as peak signal-to-noise-ratio (PSNR) or structural similarity (SSIM), is inferior to competitive methods, they delivered visually improved HR images.

Meanwhile, Gatys el al. [23] presented a very interesting approach on artistic image generation, called style transfer algorithm. Based on high-level feature maps extracted via pre-trained VGG networks [24], this algorithm synthesizes a style of an artwork to a content image, of an arbitrary photograph, while preserving the structure of the content image.

Inspired by this artistic image generation, we were wondering if we might apply this style transfer algorithm to generate the texture-enhanced image. In other words, if we can obtain a satisfactory level of HR texture image, even if the image is different from the original HR image, we can regard the obtained image as a style image and combine it with the input content image to generate a texture-enhanced HR image.

Based on this motivation, we present a novel texture enhancement framework for SISR. We call our proposed method, HR style transfer (HRST) algorithm. As shown in Fig.1, our proposed HRST algorithm provides more visually pleasing results compared to the representative state-of-the-art SISR method [19].

The remainder of this paper is organized as follows. We introduce the observation of our method in Section II. Section III describes the proposed HRST-based SR framework in detail. In Section IV, we provide experimental results with qualitative and quantitative analyses on 100 texture images. We then discuss the robustness of the proposed method and the detailed method for 4K image SR in Section V. Finally, we draw the conclusion in Section VI.

## II. OBSERVATION

It is a very challenging problem to recover the finer details of texture regions when we super-resolve at a large up-sampling factor (over $\times 4$). As shown in Fig. 1(a)-(c), the current algorithm only sharpens the lines and edges present in the interpolated LR image. However, it cannot effectively restore the high-frequency details lost by the down-sampling used to create the LR image.

To address this issue, we use the observation that texture regions tend to be comprised of visually very similar patches of various sizes. Based on this idea, we down-scale and repetitively tile the input image. This process creates a texture map we call an HR style image that is very similar in feel to the original HR image (See Fig. 2.). Here, to better correlate the image obtained via down-scaling and tiling with the original HR image, SR version of the interpolated LR image via a SISR method, namely initial HR image, may be utilized as an input image, instead of the interpolated LR image itself.

Then, we take the HR style image and combine it feature-wise with the initial HR image to generate a texture-enhanced HR image. This is different from a simple texture mapping process which just overlays one image onto a different image. Instead, our method searches both the initial HR image and the HR style image from low-level to high-level feature space for similar features. If matches are found, it strengthens them. These features are similar in terms of the correlation in feature space that is invariant to the spatial location, scale or rotation.

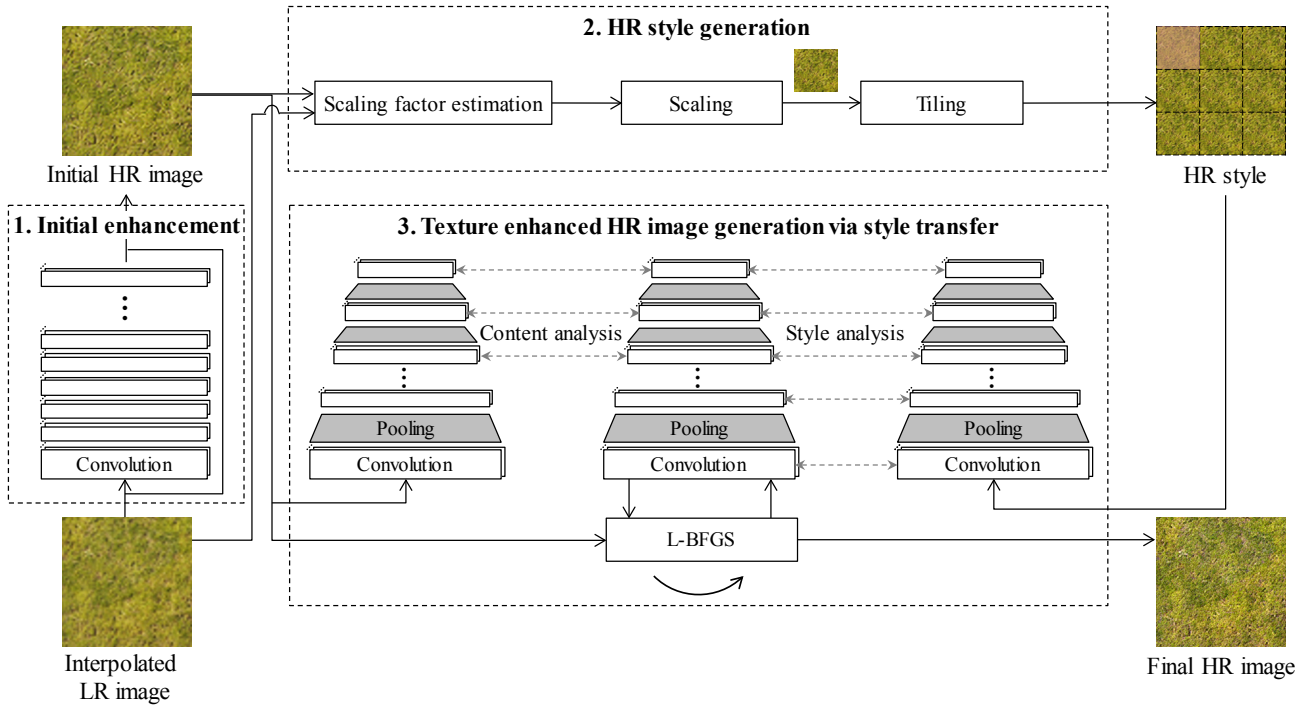

Fig. 3. Overall diagram of the proposed texture enhancement framework.

## III. PROPOSED ALGORITHM

Based on the observation detailed above, we propose a texture enhancement framework for SISR, HRST algorithm. As shown in Fig. 3, the proposed framework consists of three steps: (i) initial enhancement to generate an initial HR image from an interpolated LR image via an SISR algorithm, (ii) HR style image generation via down-scaling and tiling, and (iii) texture enhanced HR image generation by combining the HR style image with the initial HR image via a customized style transfer algorithm. In the initial enhancement step, an existing state-of-the art SISR algorithm [19] is adopted to obtain the best initial HR image.

The detail procedures based on the interpolated LR and initial HR images are presented in the following subsections.

### A. HR Style Generation

As shown in Fig. 3, the HR style image is generated by down-scaling the initial HR image and then repetitively tiling it into an image of the same size as the HR image.

In the down-scaling part, scaling factor selection is important because the scaling factor is closely related to how much the image should be improved. In general, the smaller scaling factor leads to the more HR details. However, if the selected scaling factor is enough small to exceed the detail representation of the original HR image, it may cause unnatural or overly enhanced results.

We thus introduce the process of deriving a formula that determines the proper scaling factor. In this process, we utilize 100 original HR images of 256 × 256 pixels, which are randomly cropped from 20 4K texture images, as a training image set. For each HR image, by applying the bicubic down-sampling with a factor of 0.25, we obtain a LR image.

Based on the obtained LR image, we first prepare a bicubic-interpolated LR image and an initial HR image via the SISR method [19]. Using the scale factors in the range of 0.2 to 0.9 with step size of 0.025, we generate a number of HR style images. We then obtain the final HR images corresponding to each scaling factor via a style transfer.

By comparing the final HR images with the original HR image, we find the optimal scaling factor that provides the best matched result in terms of texture detail representation. For this comparison, we adopt the mean mutual information (MMI) [25], [26] to quantify a structural complexity of the texture image. The MMI of the image was calculated as

$$\text{MMI} = \frac{4\text{ME} - \text{JE}}{3\log N_B}, \quad (1)$$

where JE represents the joint entropy among neighbor pixels, and ME represents the marginal entropy of the joint distribution. $N_B$ is the number of histogram bins. The MMI has unity value for uniform patterns and is zero for random ones.

Based on the comparison results, we visualize the tendency of the selected scaling factors, $\varphi$ depending on the MMI difference between the original and the initial HR image, $\delta$ in Fig. 4(a). It should be noted that if $\delta$ gets larger, the lower scaling factor is selected as optimum, and the relationship is nearly inverse-linear.

During the SR process, since we usually cannot use the original HR image, to estimate the $\varphi$ without using the original HR image, we further investigate the correlation between $\delta$ and $\Delta$, the MMI improvement from the interpolated LR to the initial HR image. As shown in Fig. 4(b), since the relationship can be approximately modeled linearly, we can indirectly determine the $\varphi$ based on the $\Delta$ by combining two linear relationships, this is summarized as,

$$\varphi = -4.626\Delta + 0.792. \quad (2)$$

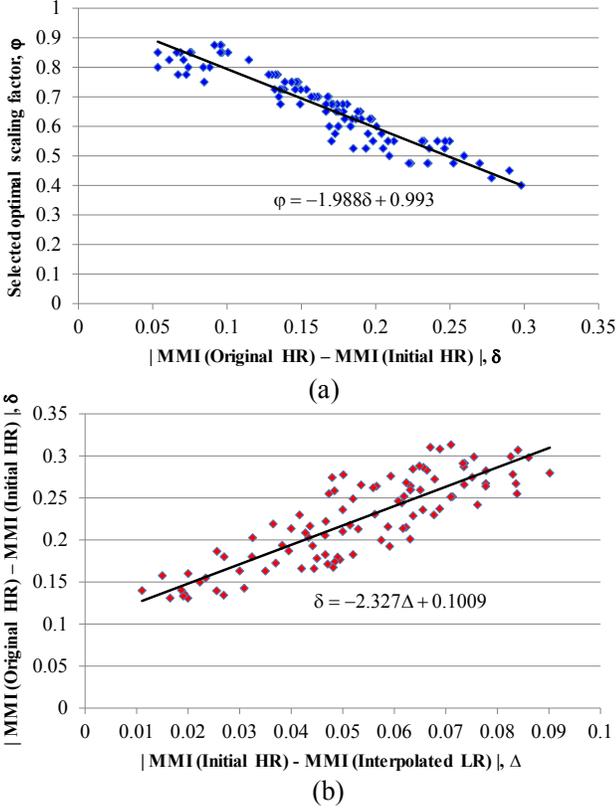

**Fig.4.** (a) Optimal scaling factors depending on the MMI deference between the original HR and the initial HR images, and (b) correlation between MMI improvement from the interpolated LR to Initial HR image and the MMI difference in (a).

To obtain the scaling factor with a unit of 0.025, (2) is reformulated as below,

$$\hat{\varphi} \equiv \left\lfloor \frac{\varphi + 0.0125}{0.025} \right\rfloor \times 0.025, \quad (3)$$

where $\hat{\varphi}$ represents the selected optimal scaling factor. Here, $\lfloor \ \rfloor$ denotes a floor operation. Note that since $\Delta$ can vary depending on the problem condition such as up-sampling factor, or initial enhancement method etc., (2) and (3) should be recalculated if the problem condition is changed.

The down-scaled image generated by the selected scaling factor is then extended to the HR image size via tiling process. In this process, to avoid an abrupt discontinuity at the tiling boundary, a boundary filtering or mirroring scheme could be applied.

### B. Texture Enhanced HR Image Generation

By using the HR style obtained in subsection III.A, we generate the final HR image with improved HR texture details, while maintaining the global characteristics of the initial HR image such as location and shape of major structures. To realize this, we adopt the style transfer algorithm [23], and customize it for better performance.

In the customized style transfer algorithm, we mainly perform two adjustments: (i) increase the number of layers used for content loss calculation, and (ii) utilize the initial HR image as an initial estimate for the final HR image.

As in [23], we performs the joint minimization of style and content losses in feature space to obtain the final HR image, $\hat{\mathbf{x}}$ by combining the HR style image with the initial HR image, which can be written as,

$$\hat{\mathbf{x}} = \min_{\mathbf{x}} \alpha L_{\text{style}}(\mathbf{x},\mathbf{s}) + \beta L_{\text{content}}(\mathbf{x},\mathbf{c}). \quad (4)$$

Here, $\alpha$ and $\beta$ are the weighting factors of style and content losses, $L_{\text{style}}$ and $L_{\text{content}}$, respectively. $\mathbf{s}$ and $\mathbf{c}$ are the HR style and the content image respectively for $L_{\text{style}}$ and $L_{\text{content}}$ calculation. $\mathbf{x}$ denotes the intermediate result image for the final HR image.

For clear description of $L_{\text{style}}$ and $L_{\text{content}}$ in feature space, we define the feature maps in the $l$-th layer from an image, $\mathbf{z}$ as $F_{\mathbf{z}}^l \in R^{C_l \times N_l}$. Here, $C_l$ is the number of feature maps and $N_l$ is the size of a feature map. $F_{\mathbf{z},ik}^l$ (or $F_{\mathbf{z},jk}^l$) denotes the $i$-th (or $j$-th) feature map at a pixel position $k$ for image $\mathbf{z}$. To extract the feature maps, we utilize the pre-trained VGG-16 network [24], which consists of 13 convolutional and 5 pooling layers.

In $L_{\text{style}}$, to analyze the style of $\mathbf{z}$ in feature space, we utilize the Gram matrix, which can measure correlations between two arbitrary feature maps at a certain layer, written as

$$G_{\mathbf{z},ij}^l = \sum_k F_{\mathbf{z},ik}^l F_{\mathbf{z},jk}^l. \quad (5)$$

To force the Gram matrix of $\mathbf{x}$, $G_{\mathbf{x},ij}^l$ similar to that of $\mathbf{s}$, $G_{\mathbf{s},ij}^l$, the energy functional for the $l$-th layer can be formulated as below,

$$E_{l,\text{style}} = \frac{1}{4C_l^2 N_l^2} \sum_{i,j} \left( G_{\mathbf{x},ij}^l - G_{\mathbf{s},ij}^l \right)^2. \quad (6)$$

Since $E_{l,\text{style}}$ compares $G_{\mathbf{x},ij}^l$ and $G_{\mathbf{s},ij}^l$, unlike $F_{\mathbf{x},ik}^l$ and $F_{\mathbf{s},ik}^l$, it can allow $F_{\mathbf{x},ik}^l$ (or $F_{\mathbf{x},jk}^l$) to be different to $F_{\mathbf{s},ik}^l$ (or $F_{\mathbf{s},jk}^l$), while making only the correlation between $F_{\mathbf{x},ik}^l$ and $F_{\mathbf{x},jk}^l$ similar to that between $F_{\mathbf{s},ik}^l$ and $F_{\mathbf{s},jk}^l$. This property can be advantageously exploited in a way that the HR style image obtained through the down-scaling and tiling process is not spatially consistent with the original HR image. Since $E_{l,\text{style}}$ does not make $F_{\mathbf{x},ik}^l$ (or $F_{\mathbf{x},jk}^l$) similar to $F_{\mathbf{s},ik}^l$ (or $F_{\mathbf{s},jk}^l$), it can implicitly prevent to transfer spatially-corresponding but unwanted image patterns of the $\mathbf{s}$ to the $\mathbf{x}$. Instead, the $E_{l,\text{style}}$ makes it possible to enhance the corresponding features of **x** if only similar correlations in feature space, irrespective of spatial location, scale or rotation, are found in both **x** and **s**.

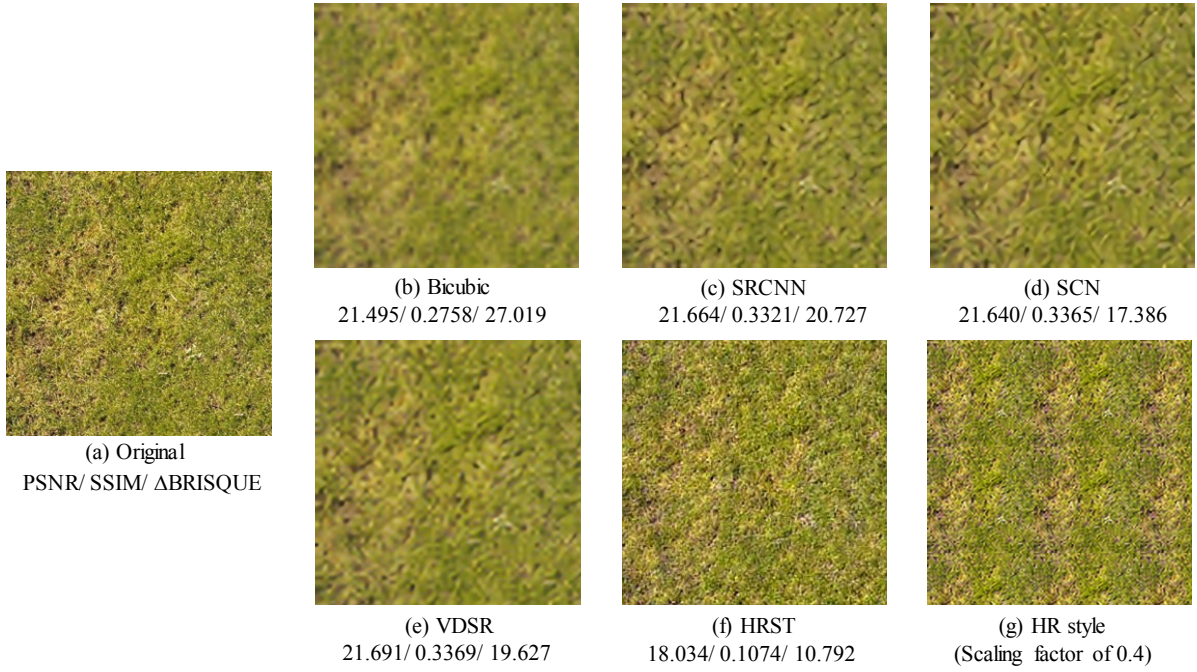

**Fig. 5.** SR results on a *grassplot* image with up-sampling factor of 4. (a) The original image, and resultant images (b-g) via (b) bicubic interpolation, (c) Dong, et al.'s [16], [17], (d) Wang, et al. [30].'s, (e) Kim et al.'s [19], (f) the proposed HRST based SR algorithm. (g) is the HR style image at a scaling factor of 0.4 used for (f).

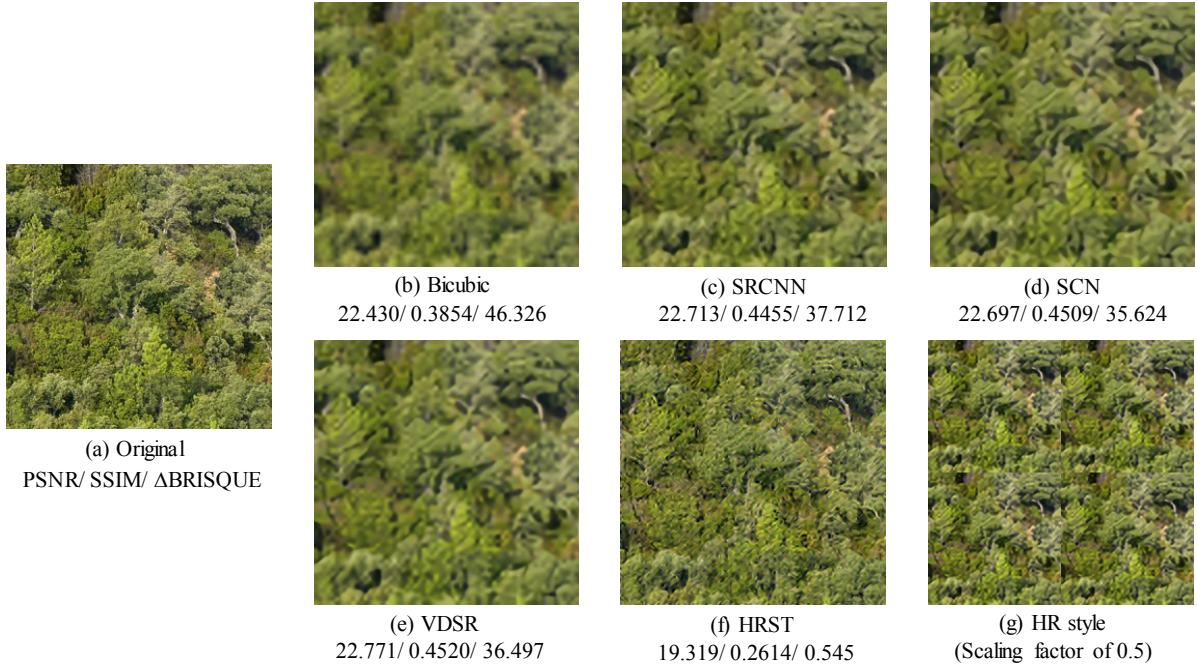

**Fig. 6.** SR results on a *forest* image with up-sampling factor of 4. (a) The original image, and resultant images (b-g) via (b) bicubic interpolation, (c) Dong, et al.'s [16], [17], (d) Wang, et al. [30].'s, (e) Kim et al.'s [19], (f) the proposed HRST based SR algorithm. (g) the HR style image at a scaling factor of 0.5 used for (f).

To reflect all the texture information of the **s** from low-level to high-level feature space, we measure the $E_{l,\text{style}}$ for each layer, and take the weighted summation of the energy functionals for $L_{\text{style}}$. This is summarized as,

$$L_{\text{style}}(\mathbf{x},\mathbf{s}) = \sum_{l \in \Omega_{\text{style}}} w_{l,\text{style}} E_{l,\text{style}}. \quad (7)$$

Here, $w_{l,\text{style}}$ is the weight factor to adjust contribution of each layer to the style loss, and $\Omega_{\text{style}}$ is set of layers for style loss calculation.

Meanwhile, to explicitly prevent that **x** becomes quite different from **c**, which is the initial HR image, the energy

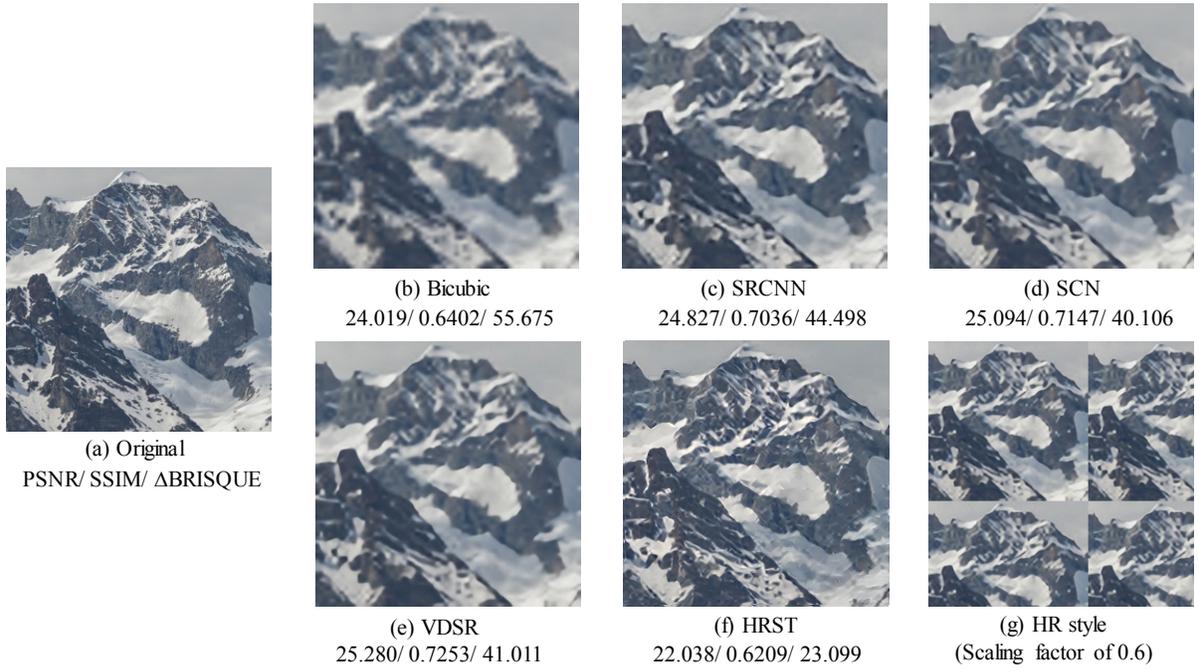

Fig. 7. SR results on a *snow mountain* image with up-sampling factor of 4. (a) The original image, and resultant images (b-g) via (b) bicubic interpolation, (c) Dong, et al.'s [16], [17], (d) Wang, et al. [30].'s, (e) Kim et al.'s [19], (f) the proposed HRST based SR algorithm. (g) the HR style image at a scaling factor of 0.6 used for (f).

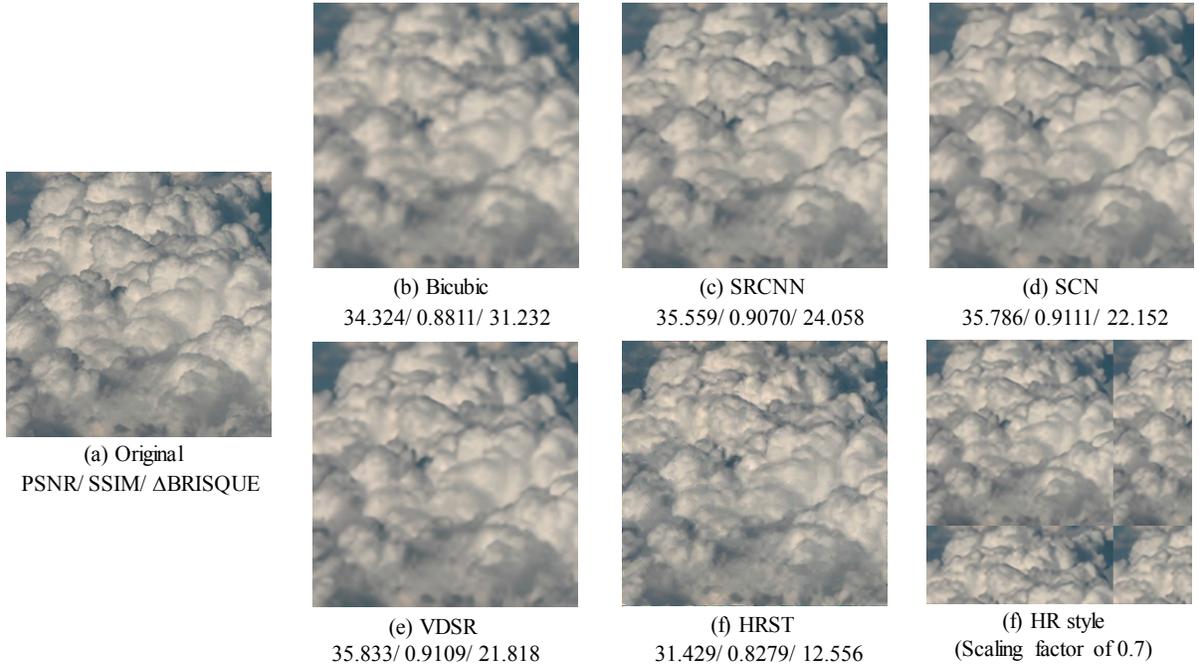

Fig. 8. SR results on a *cloud* image with up-sampling factor of 4. (a) The original image, and resultant images (b-g) via (b) bicubic interpolation, (c) Dong, et al.'s [16], [17], (d) Wang, et al. [30].'s, (e) Kim et al.'s [19], (f) the proposed HRST based SR algorithm. (g) the HR style image at a scaling factor of 0.7 used for (f).

functional for *l*-th layer can be formulated as below,

$$E_{l,\text{content}} = \frac{1}{2}\sum_{i,k}\left(F_{\mathbf{x},ik}^{l} - F_{\mathbf{c},ik}^{l}\right)^2, \quad (8)$$

Here, $\Omega_{\text{style}}$ denotes set of layers for content loss calculation.

To extend the constraints to all ranges from mid-level to high-level feature space, instead of using (8) for $L_{\text{content}}$ as in [23], we define $L_{\text{content}}$ as below,

$$L_{\text{content}}(\mathbf{x},\mathbf{h};l) = \sum_{l\in\Omega_{\text{content}}} w_{l,\text{content}} E_{l,\text{content}}. \quad (9)$$

Here, $w_{l,\text{content}}$ is the weight factor to adjust contribution of each layer to the content loss, and $\Omega_{\text{content}}$ is set of layers for content loss calculation.

To find the optimum solution, $\hat{\mathbf{x}}$ in (4), we adopt a representative gradient-based optimization, L-BFGS method [27], which provides the solution with a fast convergence. In this method, the gradient of (4) with respect to $\mathbf{x}$ can be determined based on $\partial L_{\text{style}}/\partial \mathbf{x}$ and $\partial L_{\text{content}}/\partial \mathbf{x}$ obtained via a standard error back-propagation [28].

Meanwhile, it should be emphasized that we initialize $\mathbf{x}$ with $\mathbf{c}$, instead of Gaussian random noise that is used in the existing style transfer algorithm [23]. This is an important detail not only to prevent the final HR image from being generated differently each time but also to help to preserve the original structure better. Furthermore, it leads to fast convergence.

## IV. EXPERIMENTAL RESULTS

We first describe the parameter settings used to obtain the proposed results. The results are obtained by utilizing the style feature maps on 5 convolutional layers, 'conv1', 'conv3', 'conv5', 'conv8', and 'conv11' ($w_{l,\text{style}} = 1/5$ for those layers), while utilizing the content feature maps on 3 convolutional layers, 'conv7', 'conv10', and 'conv13' ($w_{l,\text{content}} = 1/3$ for those layers). Using intermediate and higher level layers for the content feature maps helps to maintain the global and apparent structures, while allowing fine structures to be enhanced by the style feature map. The ratio $\alpha/\beta$ and the number of iterations are set to $1 \times 10^4$ and 300, respectively. To verify the performance of the proposed texture enhancement algorithm, we prepared 100 texture images. These images were cropped to a size of $256 \times 256$ pixels from 4K images. The scaling factors determined by (3) ranged from 0.4 to 0.75.

We should mention that the images in this section are not same to those used in the subsection III.A. All experiments are performed with a down-sampling factor of 4.

For performance comparison, we emphasize that the goal of this work is not to replicate the results of state-of-the-art PSNR or SSIM, but instead to demonstrate the perceptually improved visual quality. To quantify the visual improvement, we measure the difference of BRISQUE [29] metric compared with original HR image, notated as $\Delta$BRISQUE, the metric, which is known to have a high correlation with human subjective evaluation.

We compared the performance of the proposed HRST method to the bicubic-interpolation and the three different methods: the super-resolution CNN (SRCNN) [18], deep networks for super-resolution with sparse prior (SCN) [30], and very deep CNN-based super-resolution (VDSR) [19], which are currently the best performing CNN-based approaches, among the algorithms that have publically released the available code. In addition, for reference, we show the generated HR style images used for the proposed HRST.

Visual comparison of the super-resolved images is given in Figs. 5-9. For the images, the selected scaling factors were 0.4, 0.5, 0.6, and 0.7, respectively. We can note that the existing SISR algorithms poorly restore fine and detail textures, and generally provide overly-smoothed results, although they successfully enhance coarse and apparent structures. In contrast, the proposed algorithm provides finer and sharper texture representations without introducing noticeable artifacts.

TABLE I
QUANTITATIVE EVALUATION RESULTS ON AVERAGE FOR 100 TEXTURE IMAGES. THE NUMBER IN PARENTHESES DENOTES THE STANDARD DEVIATION, AND BOLD INDICATES THE BEST RESULT.

| Selected Scaling factor | Methods | PSNR | SSIM | ΔBRISQUE |
|---|---|---|---|---|
| [x0.4, x0.475] (13 images) | Bicubic | 21.63 (3.45) | 0.3787 (0.1161) | 27.63 (16.80) |
| | SRCNN | 21.91 (3.49) | 0.4371 (0.1097) | 22.75 (13.96) |
| | SCN | 21.90 (3.49) | 0.4407 (0.1104) | 23.90 (12.15) |
| | VDSR | **21.96 (3.50)** | **0.4435 (0.1103)** | 24.19 (13.23) |
| | HRST | 19.14 (3.02) | 0.2803 (0.0890) | **13.26 (12.82)** |
| [x0.5, x0.575] (23 images) | Bicubic | 20.76 (2.01) | 0.4122 (0.0796) | 39.15 (7.86) |
| | SRCNN | 21.19 (2.08) | 0.4827 (0.0810) | 31.24 (9.88) |
| | SCN | 21.19 (2.11) | 0.4871 (0.0817) | 33.47 (5.39) |
| | VDSR | **21.28 (2.13)** | **0.4919 (0.0828)** | 33.24 (10.91) |
| | HRST | 18.46 (1.75) | 0.3266 (0.0718) | **10.50 (9.76)** |
| [x0.6, x0.675] (40 images) | Bicubic | 21.29 (2.66) | 0.4551 (0.0941) | 41.14 (9.04) |
| | SRCNN | 21.82 (2.72) | 0.5320 (0.0868) | 32.23 (8.80) |
| | SCN | 21.83 (2.73) | 0.5378 (0.0882) | 37.40 (14.31) |
| | VDSR | **21.94 (2.75)** | **0.5443 (0.0888)** | 44.11 (20.02) |
| | HRST | 19.46 (2.43) | 0.4024 (0.0938) | **15.35 (8.91)** |
| [x0.7, x0.75] (24 images) | Bicubic | 23.01 (4.57) | 0.5843 (0.1314) | 36.16 (13.27) |
| | SRCNN | 24.03 (4.62) | 0.6652 (0.1106) | 31.14 (8.35) |
| | SCN | 24.18 (4.71) | 0.6752 (0.1087) | 35.36 (13.02) |
| | VDSR | **24.33 (4.62)** | **0.6852 (0.1061)** | 47.64 (23.74) |
| | HRST | 21.42 (3.56) | 0.58044 (0.1007) | **16.33 (8.31)** |

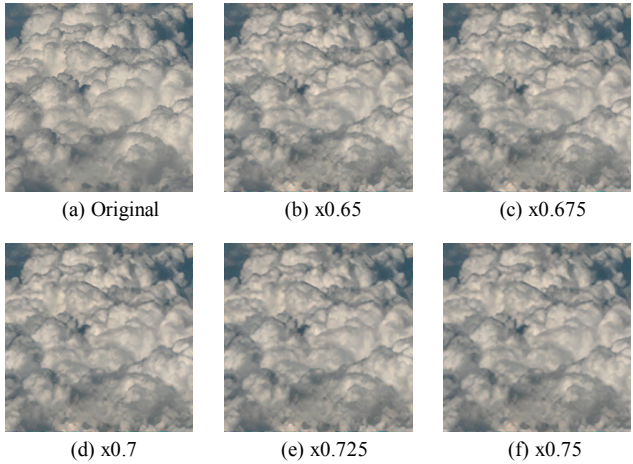

Fig. 9. Comparison of HRST results with respect to the scaling factor variation ranging from 0.65 to 0.75 with a step size of 0.025.

In addition to the subjective visual comparison, a quantitative comparison results performed using the metrics of PSNR, SSIM, and $\Delta$ BRISQUE, are summarized in Table I. We note that, the proposed method outperforms all existing SISR algorithms with a large margin in terms of $\Delta$ BRISQUE, which has an important role in perceptual image quality measurement, although it provides inferior performance in terms of PSNR and SSIM.

## V. DISCUSSIONS

The scaling factor for HR style may be critical in the proposed framework. To find the optimal scale factor, we introduced the scaling factor estimation scheme in subsection III.A. In the scheme, we approximate the correlation between the optimal scaling factor and the MMI improvement from the interpolated LR to the initial HR image with a simple linear model. This approximation may induce some estimation error; however, the scaling factor is robust to slight variations. In other words, although the final HR result may become somewhat different, it is within the observer's preference as shown in Fig. 9.

Due to the increasing use of 4K (or UHD) TV, the importance of SR for 4K images is emerging. We thus describe how to apply the proposed algorithm to real 4K images. To alleviate the hardware burden we divide a 4K image into $46 \times 25$ patches with size of $240 \times 240$. Here, we make the patches overlapped with 30% to prevent the abrupt change in the down-scaling factor. We then obtain texture enhanced HR patches independently by using the proposed HRST algorithm. Finally, we complete the final 4K image by sticking the patches with blending the overlap of each patch. As shown in Figs 10 and 11, the proposed algorithm provides vivid and realistic 4K image with finer and shaper texture details without undesirable artifacts compared to the existing SR algorithms.

As shown in the evaluation results using PSNR and SSIM in Table I, the proposed HRST needs to be further improved to achieve perfect restoration of the original image. Nonetheless, the HRST can constrain such that the generated image retains the global structures of the content image. It thereby delivers a sufficiently plausible image, although it may be somewhat different from the original image.

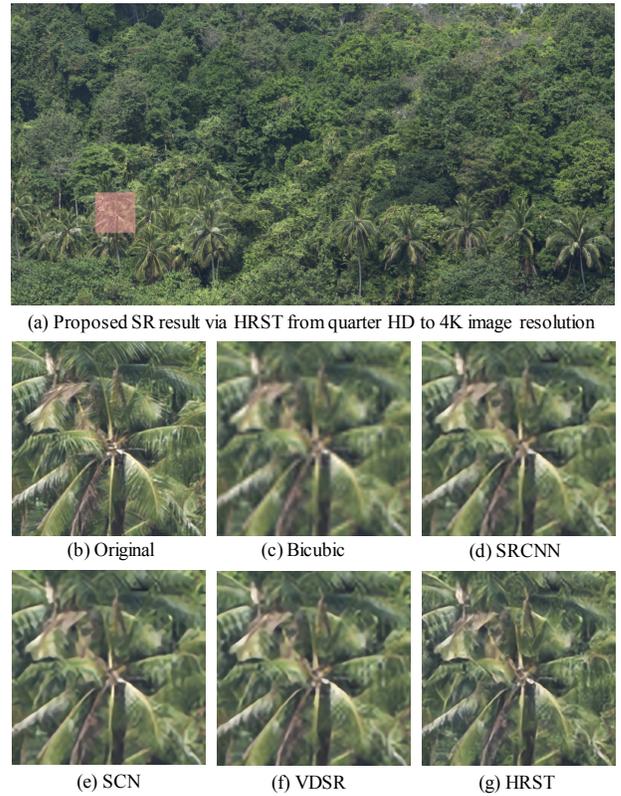

Fig. 10. SR results from quarter HD to 4K image resolution: *Palm tree* mage. The red box denotes the cropped position of the magnified images.

Since the current HRST framework is based on external optimization process, it can take up to few hours for a 4K image SR on a NVIDIA K40 GPU environment. To improve the practical applicability of the HRST, we will extend the current HRST framework to be end-to-end trainable instead of being externally optimized. We will also improve the framework to be applicable to video sequences as a future work.

## VI. CONCLUSION

In this paper, we present a novel texture enhancement framework for SISR via HR style transfer algorithm. We effectively improve the spatial resolution on the texture regions as well as edge and line regions, which is yet unresolved by existing state-of-the-art SISR algorithms. For the texture enhancement, we first obtain an initial HR image from the interpolated LR image, and then generate the HR style image from the initial HR image via down-scaling and tiling process. By properly combining semantic information of both the HR style and the initial HR images via the customized style transfer algorithm, we finally generate the texture-enhanced HR image. Experimental results demonstrate that the proposed algorithm can provide realistic and more visually pleasing SR images with finer and sharper textures, compared to the existing SR algorithms, without introducing undesirable artifacts.

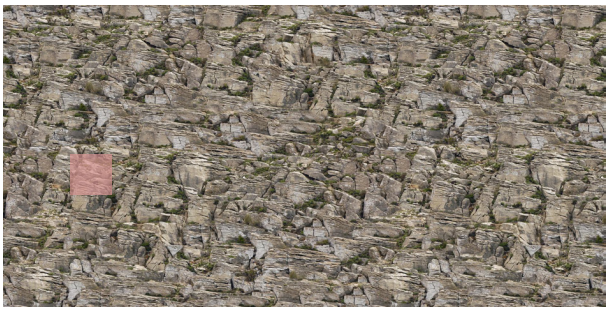

(a) Proposed SR result via HRST from quarter HD to 4K image resolution

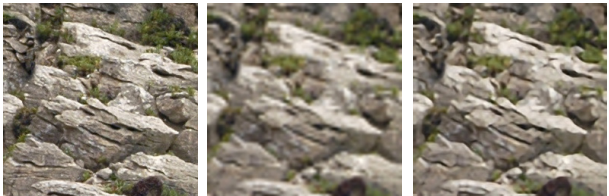

(b) Original      (c) Bicubic      (d) SRCNN

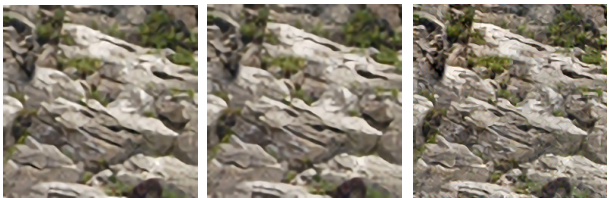

(e) SCN      (f) VDSR      (g) HRST

**Fig. 11.** SR results from quarter HD to 4K image resolution: *Rock mountain* image. The red box denotes the cropped position of the magnified images.

## BIOGRAPHIES

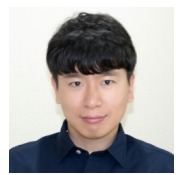

**Il Jun Ahn** received the Ph.D. degree in electrical engineering from the Korea Advanced Institute of Science and Technology (KAIST), Daejeon, in 2016. Since 2016, he has been with the Digital Media & Communications R&D Center, Samsung Electronics, where he is currently a Senior engineer. His research interests are digital image processing, video signal processing, and medical imaging.

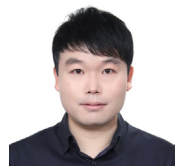

**Woo Hyun Nam** received the Ph.D. degree in electrical engineering from the Korea Advanced Institute of Science and Technology (KAIST), Daejeon, in 2013. Since 2013, he has been with the Digital Media & Communications R&D Center, Samsung Electronics, where he is currently a Senior engineer. His research interests are medical image processing and video signal processing. He is currently focused on deep learning based image/video visual quality improvement.